\documentclass[11pt,a4paper]{article}
\usepackage{times}
\usepackage{acl}

\usepackage{url}

\usepackage{amsmath}
\usepackage{amsthm}
\usepackage{amssymb}
\usepackage{amsfonts}
\usepackage{microtype}
\usepackage{xfrac}
\usepackage{xspace}
\usepackage[belowskip=-10pt,aboveskip=10pt]{caption}
\usepackage{subcaption}
\usepackage{enumitem}
\setlist[itemize]{topsep=-0.1cm}

\newtheoremstyle{named}{}{}{\itshape}{}{\bfseries}{.}{.5em}{\thmnote{#3} #1}
\theoremstyle{named}
\newtheorem*{namedtheorem}{Hypothesis}

\usepackage{cleveref}
\crefname{section}{\S}{\S\S}
\Crefname{section}{\S}{\S\S}
\crefname{table}{Tab.}{}
\crefname{figure}{Fig.}{}
\crefname{algorithm}{Algorithm}{}
\crefname{equation}{Eq.}{}
\crefname{appendix}{App.}{}
\crefformat{section}{\S#2#1#3}  

\newcommand{\heads}{\texttt{H}}
\newcommand{\tails}{\texttt{T}}
\newcommand{\pheads}{p(X\!=\!\heads)}
\newcommand{\ptails}{p(X\!=\!\tails)}
\newcommand{\ent}{\mathrm{H}}

\newcommand{\defn}[1]{\textbf{#1}}
\newcommand{\xx}{\mathbf{x}}

\newcommand{\yy}{\mathbf{y}}

\newcommand{\calY}{\mathcal{Y}}
\newcommand{\calC}{\mathcal{C}}

\newcommand{\defeq}[0]{\mathrel{\stackrel{\textnormal{\tiny def}}{=}}}

\newcommand{\vtheta}{{\boldsymbol \theta}}

\newcommand{\vocab}{\mathcal{V}}

\newcommand{\eos}{\textsc{eos}\xspace}
\newcommand{\bos}{\textsc{bos}\xspace}
\newcommand{\emp}{p}
\newcommand{\model}{q}
\newcommand{\information}{\textsc{i}}

\usepackage[disable]{todonotes}
\makeatletter
\newcommand*\iftodonotes{\if@todonotes@disabled\expandafter\@secondoftwo\else\expandafter\@firstoftwo\fi}  
\makeatother

\usepackage{emoji}

\newcommand{\ethz}{\emoji[openmoji]{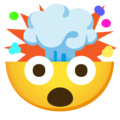}}
\newcommand{\ucambridge}{\emoji[openmoji]{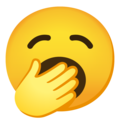}}
\usepackage{inconsolata}

\usepackage{color}

\title{On the probability--quality paradox in language generation}

\author{
 Clara Meister$^{\ethz}$~\;~ Gian Wiher$^{\ethz}$~\;~Tiago Pimentel$^{\ucambridge}$~\;~Ryan Cotterell$^{\ethz}$ \\
   $^{\ethz}$ETH Zürich~\;~ $^{\ucambridge}$University of Cambridge\\
  \texttt{\href{mailto:clara.meister@inf.ethz.ch}{clara.meister@inf.ethz.ch}}~\;~\texttt{\href{mailto:gian.wiher@inf.ethz.ch}{gian.wiher@inf.ethz.ch}} \\
   \texttt{\href{mailto:tp472@cam.ac.uk}{tp472@cam.ac.uk}}~\;~\texttt{\href{mailto:ryan.cotterell@inf.ethz.ch}{ryan.cotterell@inf.ethz.ch}}
}

\date{}

\begin{document}
\maketitle
\begin{abstract}
When generating natural language from neural probabilistic models, high probability does not always coincide with high quality: 
It has often been observed that mode-seeking decoding methods, i.e., those that produce high-probability text under the model, lead to unnatural language. On the other hand, the  lower-probability text generated by stochastic methods is perceived as  more human-like. 
In this note, we offer an explanation for this phenomenon by analyzing language generation through an information-theoretic lens. 
Specifically, we posit that human-like language should contain an amount of information (quantified as negative log-probability) that is close to the entropy of the distribution over natural strings. Further, we posit that language with substantially more (or less) information is undesirable. 
We provide preliminary empirical evidence in favor of this hypothesis; quality ratings of both human and machine-generated text---covering multiple tasks and common decoding strategies---suggest high-quality text has an information content significantly closer to the entropy than we would expect by chance.\looseness=-1
\end{abstract}

\section{Introduction} 
Today's probabilistic neural language models are often trained on millions---if not billions---of lines of human text; thus, at least at an intuitive level, we would expect high-probability generations to be human-like. 
Yet the high-quality\footnote{We assume that ``human-like'' is a (necessary but not sufficient) prerequisite for ``high-quality'' in the context of natural language strings.} texts these models have become famous for producing  \cite{NEURIPS2020_GPT3,clark-etal-2021-thats} are usually not 
those assigned the highest probability by the model \cite{fan_hierarchical_2018,holtzman_curious_2020,BasuRKV2021,delucia_decoding_2020}.
Rather, the relationship between probability and quality appears to have an inflection point,\footnote{The inflection point is empirically demonstrated in our \cref{app:exp} or in Fig. 1 of \citet{zhang_trading_2020}.} i.e., quality and probability are positively correlated only until a certain threshold, after which the correlation becomes negative.
While the existence of such a trend has received informal explanations (see, e.g.,  \citet{ippolito_comparison_2019} and \citet{zhang_trading_2020} for a qualitative discussion about the trade-off between diversity and quality), it lacks a more fundamental understanding.
Why does the lower probability text produced by stochastic decoding methods---such as nucleus or top-$k$ sampling---outperform text generated using probability-maximizing approaches? 
In this note, we take an information-theoretic approach in an attempt to answer this question.\looseness=-1

In information theory, probability has another interpretation: its negative log quantifies \defn{information content}. 
In the context of natural language, the notion of information content is intuitive; humans use strings as a means to convey information.
Further, less predictable text, i.e., text which would be harder for us to anticipate, conveys \emph{more} information. 
If we assume that the goal of human communication is to transmit messages efficiently and reliably \cite{gibson2019efficiency}, we may predict that these strings' information content should concentrate inside a specific interval.
At one extreme, strings with more-than-expected information may be hard to process, and thus ought to be disfavored when producing language.\footnote{Many works in psycholinguistics have shown a direct relationship between information content and processing effort \cite[\textit{inter alia}]{smith2013effect,wilcox2020predictive}.} 
At the other extreme, low-information strings may be seen as boring and uninformative.\looseness=-1

Collectively, these concepts lead us to propose the \defn{expected information hypothesis}: Text perceived as human-like should have an information content within a small interval around the expected information---i.e., the entropy---of natural language strings. 
Such a hypothesis offers an intuitive explanation for the trends observed in natural language generation (NLG), i.e., why desirable text seems to exist not always at the high end of the probability spectrum but around a certain inflection point.\footnote{Similar ideas have been used to improve language models and language generation before \cite{meister_if_2021,wei-etal-2021-cognitive}. } 
Moreover, it also gives us a \emph{testable} hypothesis: given a language generation model $\model$ whose entropy we can empirically estimate, we can evaluate whether high-quality text indeed has an information content that falls within an interval around this quantity.\looseness=-1



To test our hypothesis, we perform an analysis comparing human and model-generated text, investigating multiple common decoding strategies and NLG tasks. 
Specifically, our analysis focuses exclusively on English text.
We indeed observe that the information content of highly ranked text (as judged by humans) often 
falls within a standard deviation of model entropy; there is statistically significant evidence that this is not due to chance.
Further, the best-performing decoding methods appear to select strings with an information content within this interval.
We take these observations as empirical support for our hypothesis, helping to explain the probability--quality paradox observed in language generation.\looseness=-1

\section{Probabilistic Language Generators}\label{sec:gen}
In this work, we focus on probabilistic models for language generation tasks. 
Formally, these models are probability distributions $\model$ over natural language strings $\yy \in \calY$, where $\calY$ is the (countably infinite) set consisting of all possible strings that can be constructed from a set vocabulary $\vocab$:\looseness=-1
\begin{equation}
    \calY \defeq \{ \bos \circ \mathbf{v} \circ \eos \mid \mathbf{v} \in \vocab^* \}
\end{equation}
Here, $\bos$ and $\eos$ stand for special reserved beginning- and end-of-string tokens, respectively, and $\vocab^*$ denotes the Kleene closure of $\vocab$. In practice, we limit the set of strings we consider to $\calY_N \subset \calY$ for some maximum sequence length $N$.\looseness=-1

Note that $\model$ may be a conditional model. 
For instance, we may model $\model(\cdot \!\mid\! \xx)$ where $\xx$ is an input text, as in the case of machine translation, or an input image, as in the case of image captioning.
However, for notational brevity, we omit this explicit dependence in most of our subsequent analyses.
In order to estimate $\model$, it is standard practice to maximize the log-probability of a training corpus $\calC$ under the model with respect to the model's parameters $\vtheta$. 
This is equivalent to minimizing its negative log-probability:\looseness=-1
\begin{align}\label{eq:nll}
        L(\vtheta;\calC) = -\sum_{\yy \in \calC}\log \model(\yy)
\end{align}
There are many different decision rules one can employ for generating natural language strings from a model $q$; such sets of rules are generally referred to as decoding strategies; see \citet{wiher-decoding} for an in-depth review.
Given the probabilistic nature of the models we consider, an intuitive strategy for decoding would be to choose the string with the highest probability under $q$, an approach referred to as maximum-a-posteriori (MAP) decoding.\footnote{Note that MAP decoding is somewhat of a misnomer since we are not maximizing over a Bayesian posterior. Nonetheless, the term has become commonplace in the language generation literature.}
Yet recent research has shown that solutions to MAP decoding---or, even more generally, to heuristic mode-seeking methods such as beam search---are often not high-quality, even in state-of-the-art NLG models. 
For example, in the domain of machine translation, the most probable string under the model is often the empty string \cite{stahlberg_nmt_2019}. Similarly, in the domain of open-ended generation, mode-seeking methods produce dull and generic text \cite{holtzman_curious_2020}.\looseness=-1

Where maximization has failed, authors have turned to stochastic methods, taking random samples from $\model$. 
While the resulting text is often assigned much lower probability than the mode, it can be qualitatively much better. 
This peculiarity has puzzled the language generation community for the last few years, with only qualitative intuitions being offered as explanation. This paper in turn offers a quantitative explanation.\looseness=-1


\section{Language as Communication}\label{sec:information-theory}

While many aspects of natural language may not perfectly adhere to \citeauthor{shannon1948mathematical}'s mathematical theory of communication, 
there are several characteristics of human language that \emph{can} fruitfully be described using an information-theoretic framework.\footnote{A large body of work has explored the extent to which attributes of human languages---such as word lengths or phoneme distributions---can be explained as information-theoretic design features \cite{gibson2019efficiency}. Surprisal theory, for instance, directly relates human language processing difficulty to information content \citep{hale2001probabilistic}.}
Here we employ this framework for explaining recent phenomena observed in probabilistic NLG.

\subsection{Measuring Information}
We can precisely compute the information content of a string given the \emph{true} (perhaps conditional) probability distribution $\emp$ over natural language strings.
Fortunately, this is the exact distribution our language generation models in \cref{sec:gen} are trained to approximate.\footnote{To see this, recall that minimizing the objective in \cref{eq:nll} is (up to an additive constant) equivalent to minimizing the Kullback--Leibler divergence---an information-theoretic quantity that measures the amount of information lost when approximating one probability distribution with another---between the empirical distribution $\emp$ and our model $\model$.\looseness=-1} 
Assuming $\model$ approximates $\emp$ well (as quantified by metrics such as perplexity), we may thus use it to estimate such attributes of natural language strings. 
In this work, we will measure the amount of information a specific realization $\yy$ contains, which we denote $\information(\yy) \defeq -\log \model(\yy)$, as well as the \emph{expected} amount of information a random $\yy \in \calY_N$ drawn from $\model$ contains, also termed the entropy of $\model$:\looseness=-1
\begin{equation}\label{eq:ent}
    \mathbb{E}_\model\left[\information(\yy)\right] =\ent(\model) = -\!\!\sum_{\yy \in \calY_N} \model(\yy) \log \model(\yy)
\end{equation}
Note that \citet[Theorem 2]{pimentel-etal-2021-non} prove that, as long as the probability of $\eos$ under $\model$ is bounded below by some $\epsilon > 0$, then the entropy of $\model$ is finite. In our case we restrict $q$ to a finite subset $\calY_N$ of $\calY$, which also implies that \cref{eq:ent} is finite.\looseness=-1


\subsection{The Expected Information Hypothesis}

Language is used as a means for transferring information. This property of language has in fact motivated several theories of language evolution; many have posited, for instance, that natural language has developed to optimize for reliable and efficient data communication, subject to cognitive resources \cite{zipf1949human,hockett1960origin,hawkins2004efficiency,piantadosi2011word}.  
The above theories arguably imply that humans tend to produce natural language strings with a certain amount of information; they also imply that, on the receiving end of communication, humans would expect similar strings.
We argue that this amount is intuitively close to the language's entropy, i.e., close to the average string's information content.
\looseness=-1

\begin{namedtheorem}[Expected Information]
Text perceived as human-like typically encodes an amount of information close to the expected information content of natural language strings, i.e., in the interval $[\ent(\emp)-\varepsilon,\, \ent(\emp)+\varepsilon]$ for a natural language string distribution $p$ and some $\varepsilon$.\footnote{While we do not offer a concrete explanation of why distributions over natural language strings have a particular entropy, we posit that it is determined by cognitive constraints, as observed with other phenomena in natural language \cite{coupe,pimentel-etal-2021-surprisal}.} Text that falls outside of this region is likely perceived as unnatural.\looseness=-1  
\end{namedtheorem}
\noindent This viewpoint can be applied to the problem of decoding neural text generators. In the context of a model $\model$ of the distribution $\emp$, this implies that---when $\model$ is a good approximation---human-like text should typically have a negative log-probability close to the entropy of $\model$. In \cref{sec:exps}, we provide empirical evidence for this hypothesis.\looseness=-1

\paragraph{Relationship to the typical set.}
The set of strings that we discuss has an intuitive relationship to the typical set \cite{shannon1948mathematical}, an information-theoretic concept defined for stationary ergodic stochastic processes. However, generation from standard neural probabilistic language models cannot be framed as such a process.\footnote{Specifically, most neural language models are neither stationary (due to their ability to encode arbitrarily long sequences; \citealt{welleck_consistency_2020}) nor ergodic (because of the absorbing nature of the \eos state). This implies that we cannot guarantee the existence of an entropy rate, which is necessary to define the typical set.} While we cannot  utilize the formal mathematical underpinnings of typicality, the connection can still be useful for understanding why strings with a given information content exhibit certain characteristics. An overview of the concept is in \cref{app:typical} for the interested reader; also see \citet{dieleman2020typicality} for further insights on typicality in the context of generative models.\looseness=-1



\section{Experiments}\label{sec:exps}
Our experiments present an analysis of the distribution of information content in text generated by both humans and probabilistic models. Specifically, we look at the relationship between information content and quality---as measured by human judgments. We perform experiments on two natural language generation tasks: abstractive summarization and story generation. We present the results for story generation here, while the results for summarization can be found in \cref{app:exp} due to space constraints. A recreation of the probability versus quality plots of \citet{zhang_trading_2020} can also be found in \cref{app:exp}.

We use the following Monte Carlo estimator for the entropy, i.e., expected information content, of our model $\model$:
\begin{align}
  \widehat\ent(\model) =\frac{1}{M} \sum_{m=1}^M -\log \model(\yy^{(m)})
\end{align}
where we sample $\yy^{(m)} \overset{\text{i.i.d.}}{\sim} \model$. Algorithmically, taking these samples may be done in linear time using ancestral sampling. All computations are performed with the test sets of respective datasets. 
Note that for both abstractive summarization and story generation, where we condition on some input $\xx$, we must compute the \emph{conditional} entropy for each input, i.e., using $\model(\cdot \mid \xx)$ instead of $\model(\cdot)$. For each $\xx$, we take $M=100$ to estimate $\widehat\ent(\model(\cdot \mid \xx))$.  

\subsection{Setup}
\paragraph{Models and Data.} 
We only conduct experiments on the English language.
For story generation, we fine-tune GPT-2 (medium)  \cite{radford_language_nodate} (checkpoint made available by OpenAI) on the \textsc{WritingPrompts} dataset \cite{fan_hierarchical_2018}. For abstractive summarization, we use BART \cite{lewis_bart_2019}, fine-tuned on the \textsc{CNN/Dailymail} dataset \cite{nallapati-etal-2016-abstractive}. We rely on the open-sourced code-base from the \href{https://huggingface.co/}{HuggingFace} framework \cite{wolf-etal-2020-transformers} for reproducibility.\looseness=-1

 \begin{figure}
    \centering
    \includegraphics[width=\linewidth]{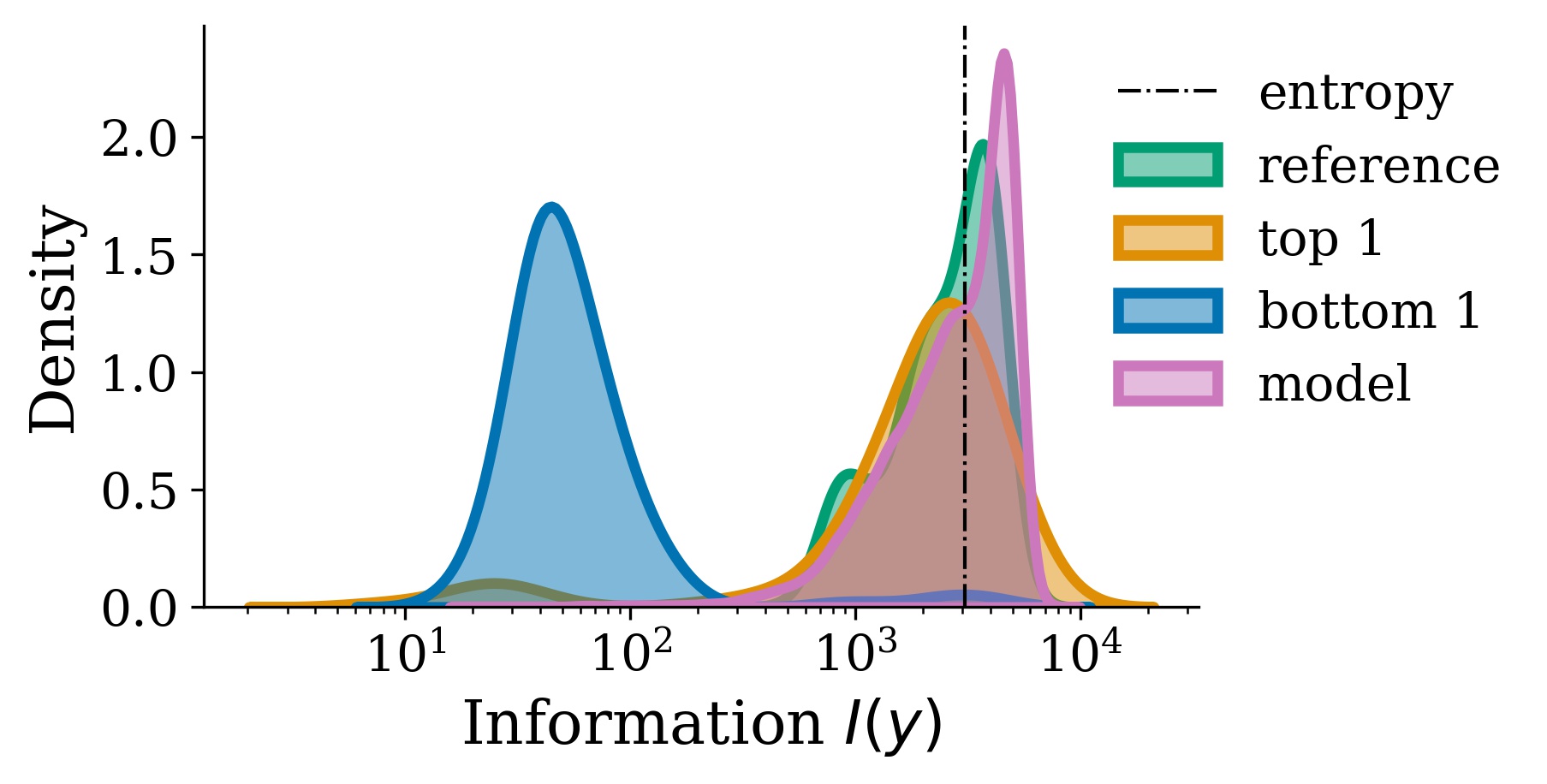}
    \caption{The distribution over information $\information(\yy)$ values of:  \textsc{model}, the model, as estimated using samples from $\model$; \textsc{reference}, the reference strings; \textsc{top 1}  and \textsc{bottom 1}, model-generated strings ranked first and  last (respectively) among all decoding strategies by human annotators. The latter 3 are all w.r.t. a held-out test set. Same graph is reproduced for individual decoding strategies in \cref{app:exp}. }
    \label{fig:dists}
\end{figure}

\paragraph{Decoding Strategies.} We explore text generated according to a number of different decoding strategies. Unless otherwise stated, we use the implementation provided by Hugging Face for each of the decoding algorithms. 
Along with standard ancestral sampling, we experiment with the following six decoding strategies:
\begin{itemize}[noitemsep]
    \item \textbf{greedy search};
    \item \textbf{beam search} with beam sizes $k=5$ and $k=10$;
    \item \textbf{diverse beam search} \cite{vijayakumar_diverse_2018} with Hamming distance as a dissimilarity function and $\lambda=0.7$ and $G=k=5$;\footnote{The choice of dissimilarity function and hyperparameters ($\lambda,G,k$) is based on the recommendations from the original work.} 
    \item \textbf{ancestral sampling};
    \item \textbf{top-$k$ sampling} \cite{fan_hierarchical_2018} with $k=30$;
    \item \textbf{nucleus sampling} \cite{holtzman_curious_2020} with $p=0.85$;\footnote{This choice is based on experiments in \cite{delucia_decoding_2020} that suggest a parameter range $p \in [0.7, 0.9]$.}
    \item \textbf{minimum Bayes risk decoding} (MBR; \citealt{eikema_is_2020})\footnote{We use the \href{https://github.com/Roxot/mbr-nmt}{github.com/Roxot/mbr-nmt} framework.} with $32$ Monte Carlo samples\footnote{The number of Monte Carlo samples was chosen based on the batch size constraint.} from $\model$ and BEER \cite{stanojevic_fitting_2014} as the utility function.\looseness=-1
\end{itemize}    
\paragraph{Human Evaluations.}  We  use the  \href{https://www.prolific.co/}{\emph{prolific}} platform to obtain human judgments of text quality (according to 2 criteria per task) from 5 different annotators on 200 examples per decoding strategy--per task. This gives us a total of $>3000$ annotated examples. We largely follow the guidelines recommended by \citet{eval_guide} in setting up our evaluations: For abstractive summarization, we ask annotators to rate \emph{quality} and \emph{accuracy} while for story generation, annotators rate \emph{fluency} and \emph{naturalness}.
More details on our setup can be found in \cref{app:prolific}.\looseness=-1

\begin{figure}
    \centering
    \includegraphics[width=\linewidth]{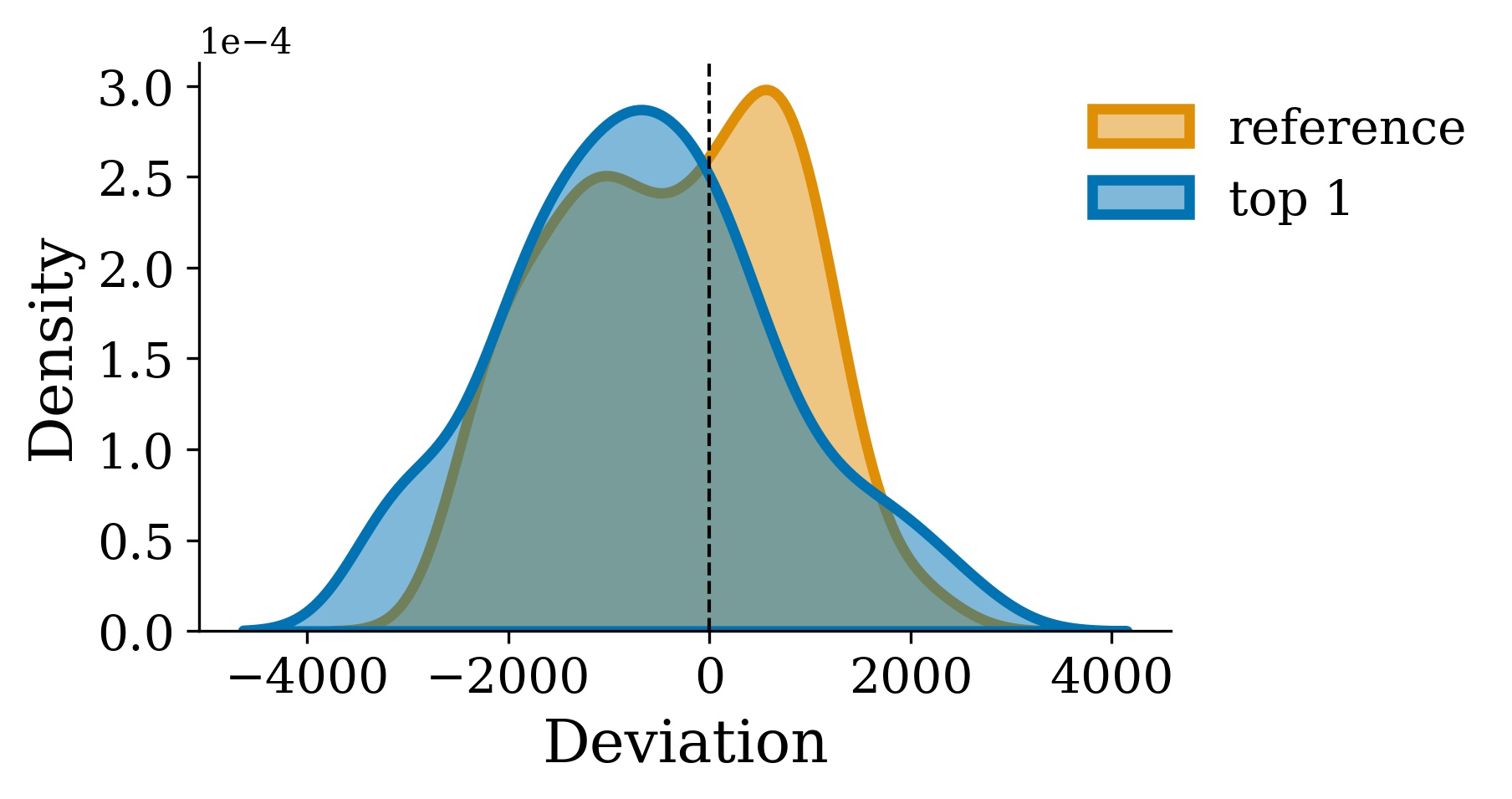}
    \caption{The distribution of the difference in total information content for (1) test-set references and (2) top-ranked model-generated strings from the (conditional) entropy of the model from which they were generated.}
    \label{fig:ent-diff}
\end{figure}

\subsection{Results}
In \cref{fig:dists}, we plot the distribution of information content assigned by $\model$ to four different sets of strings: our reference (human-generated) text, the top and bottom ranked (according to human annotators) strings generated from $\model$ via our different decoding strategies,\footnote{Specifically, for each input, we generate a single string according to each decoding strategy. We then rank these strings according to scores from human annotators.} and strings sampled i.i.d. from $\model$. Note that the latter should represent the distribution of negative log-probabilities assigned to strings by the model. 
We see that both the references and the top-ranked model-generated strings---both of which we assume are of relatively high quality---contain an amount of information clustered around the (estimated) model entropy. On the other hand, the distribution of the information content of poorly rated strings is skewed towards much lower values.
The same trends hold when looking at information normalized by string length, i.e., $\information(\yy)/|\yy|$ (see \cref{app:exp}), demonstrating these trends are not purely an artifact of string length. We note that in our human evaluations, the reference string was ranked first in $47\%$ of cases and it was tied for first in an additional $16\%$ of the cases. 
This suggests that the quality of the reference strings is on par with---if not higher than---the set of ``top 1'' model-generated strings.

\cref{fig:ent-diff} shows the distribution of deviations of strings' information content from the model entropy;\footnote{Note that this is not simply \cref{fig:dists} shifted by a constant, as deviations are computed w.r.t. input-dependent conditional entropy estimates, i.e., $\widehat\ent(\model(\cdot \mid \xx))$.} results are shown for both reference strings and top-ranked model-generated strings. Because these values are distributed quite evenly around 0, we take this as additional evidence that high-quality text usually has information content close to $\ent(\model)$. Further, the shapes of these curves motivate us to perform our next set of tests using $\varepsilon=\sigma$, the standard deviation of information values under $\model$.\footnote{Similarly to our estimation of $\ent(\model)$ in \cref{eq:ent}, $\sigma$ can be estimated from the distribution of values of $\information(\yy)$ sampled from the model.}\looseness=-1
\begin{figure}
    \centering
    \includegraphics[width=\linewidth]{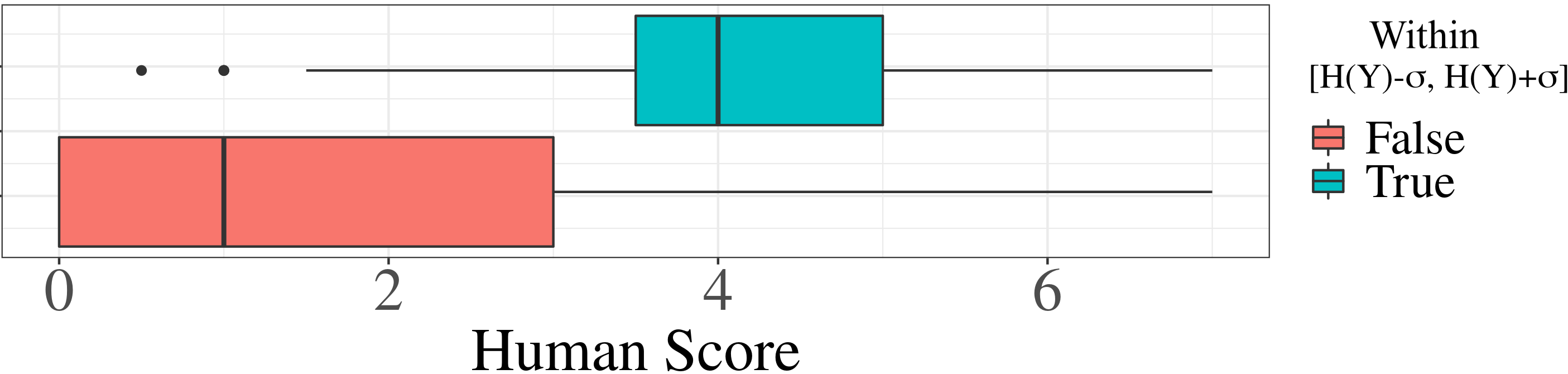}
    \caption{Human scores for strings (including both reference text and model-generated text) within 1 std of model entropy and outside of this interval. There is a statistically significant difference in means ($p < 0.001$).}
    \label{fig:box}
\end{figure}

We employ statistical hypothesis testing to see if the percentage of high-quality strings whose information content falls in the interval $[\ent(\model)-\sigma,\, \ent(\model)+\sigma]$ is greater than chance. 
For each input $\xx$ (i.e., either a story prompt or article), we compute the information content of the reference and top-3 human-ranked strings. 
We then compute the percentage of items (among these four) that fall within $[\ent(\model(\cdot \mid \xx))-\sigma,\, \ent(\model(\cdot \mid \xx))+\sigma]$. 
We compare this percentage to the percentage of strings sampled directly from $\model(\cdot \mid \xx)$ that falls within this interval. 
The former should (in expectation) be greater than the latter if the probability of high-quality strings having information content within this interval is greater than chance. 
Specifically, we test this using a paired, unequal-variance $t$-test, where samples with the same input are paired. 
At significance level $\alpha = 0.01$, we reject our null hypothesis---i.e., we reject that the percentage of highly rated strings (reference plus top-3 human-ranked strings) that fall within this interval is equal to (or less than) what we should expect by chance.
Further, using a simple unpaired $t$-test, we find that the mean human score of strings (across all decoding strategies) within this region is significantly higher than those outside of this region. This characteristic is visualized in \cref{fig:box}, where we plot the distributions of human quality ratings for strings inside and outside of this interval. We include a version of \cref{fig:box} further broken down by whether strings fall \emph{above} or \emph{below} this interval in \cref{app:exp}. \looseness=-1

Additional plots reinforcing these observations can be found in \cref{app:exp}.
Also see \citet{meister+al.pre22} for follow-up experiments to this work.

\section{Conclusion}
In this work, we present the \defn{expected information hypothesis}, which states that human-like strings typically have negative log-probability close to the expected information content of the probabilistic model from which they were generated. We use this hypothesis to explain why high-quality text seems to exist not necessarily at the high end of the probability spectrum but, rather, close to the entropy of the model. We provide empirical evidence in support of our hypothesis in an analysis of both human and machine-generated text, demonstrating that, overwhelmingly, high-quality text indeed has information content in the proposed region.\looseness=-1

\section*{Ethics Statement}
In order to complete our human evaluation, we used a crowdsourcing platform. 
For each task, we estimated the amount of time we expected the task to take and made sure that the crowdworkers would be paid (at minimum) a wage of \$15 per hour. 
A further ethical consideration of this work is in the context of the use of language models for text generation. Language models have been used for the generation of malicious text, e.g., fake news and triggering content. 
The results in this work may provide insights for those using language models for such purposes as to how generations can be chosen to seem more ``human-like.'' 

\section*{Acknowledgments}
The authors would like to thank the anonymous reviewers for their helpful recommendations, as well as Sander Dieleman for feedback on a preliminary draft of this paper.


\bibliography{acl}
\bibliographystyle{acl_natbib}

\appendix
\onecolumn
\newpage
\clearpage

\section{The Typical Set}\label{app:typical}
Let us imagine flipping $N$ biased coins; specifically, let $X \sim p$ be an indicator random variable that takes values $\heads$ and $\tails$. 
Take $\pheads = 0.6$ and $\ptails = 0.4$. Flipping $N$ biased coins is then equivalent to taking $N$ i.i.d. samples $x_n \sim p$.  For reasonably large $N$, what might you expect the sequence $x_1, \ldots, x_N$ to look like?  Few people would answer ``all heads,'' even though this is technically the highest probability sequence. Rather, intuition tells you: an expected sequence would be one comprised of approximately 60\% heads and 40\% tails. 

The samples that fall into the latter category have a distinctive characteristic: they contain a near-average amount of information w.r.t the support of the distribution over $X_1,\ldots, X_N$, where the information content of a realization $x_1,\ldots, x_N$ is defined as its negative log-probability. 
More formally, the (weakly) \defn{$(\varepsilon,N)$-typical set} $A^{(N)}_\varepsilon$ for a chosen $\varepsilon >0$ is the set of assignments $x_1,\ldots, x_N$ to random variables $\overrightarrow{X} = X_1, \ldots, X_N$ such that
\begin{align}\label{eq:typical}
    2^{-N( \ent(p)+\varepsilon)} 
   & \leqslant p(x_1, \ldots, x_N) \leqslant
    2^{-N( \ent(p) - \varepsilon)} \nonumber
\end{align}
where $\ent(p) \defeq -\sum_{x} p(x) \log p(x)$ is the entropy---or equivalently, the expected value of the information content---of the random variable $X$. 
Under this definition we can prove that, for every $\varepsilon > 0$, there exists an $N_0$ such that for all $N > N_0$, we have that the $(\varepsilon,N)$-typical set contains at least $(1-\varepsilon)$ 
of the probability mass of the joint distribution over $\overrightarrow{X}$.
The concept of the typical set also generalizes to stochastic processes when we can actually compute their average information rate---or equivalently, their entropy rate.\looseness=-1

\section{Experimental Design}\label{app:exp}

\subsection{Human Evaluations}\label{app:prolific}
For story generation and abstractive summarization, the raters are first presented with a news article/prompt. Next, they are presented, in random order, with the corresponding reference and the summaries/stories generated by different decoders. 
For each of two rating criteria, a score from 0 to 7 is assigned. For story generation the criteria are \textsc{fluency} and  \textsc{naturalness} while for abstractive summarization \textsc{quality} and \textsc{accuracy} are used. We provide the following short descriptions of the criteria to the raters: 
\begin{itemize}
    \item[] \textsc{Fluency}: How fluent is the English text?
    \item[] \textsc{Naturalness}: Does the text seem to be natural English text?
    \item[] \textsc{Quality}: How high is the overall quality of the text?
    \item[] \textsc{Accuracy}: How well does the summary summarize the article?
\end{itemize}\vspace{.25cm}
After we obtain the ratings, we reject ratings that have not been filled out with care. Specifically,  a rater is rejected if he assigns high scores to multiple examples that do not fulfill the specified criteria at all. If a rater has been rejected, we obtain a fresh set of ratings from a new rater.

\section{Additional Figures}
We provide several additional results, looking further into the relationship between text information content and perceived quality. We see that in general, the distribution of information content of reference strings is quite close to that of the model. While the distribution of information content of top 1 ranked strings is also closer to the model distribution than many of the individual decoding strategies, the overlap is not as high as for reference strings.
\newpage
\clearpage
\twocolumn
\begin{figure}
    \centering
    \includegraphics[width=\linewidth]{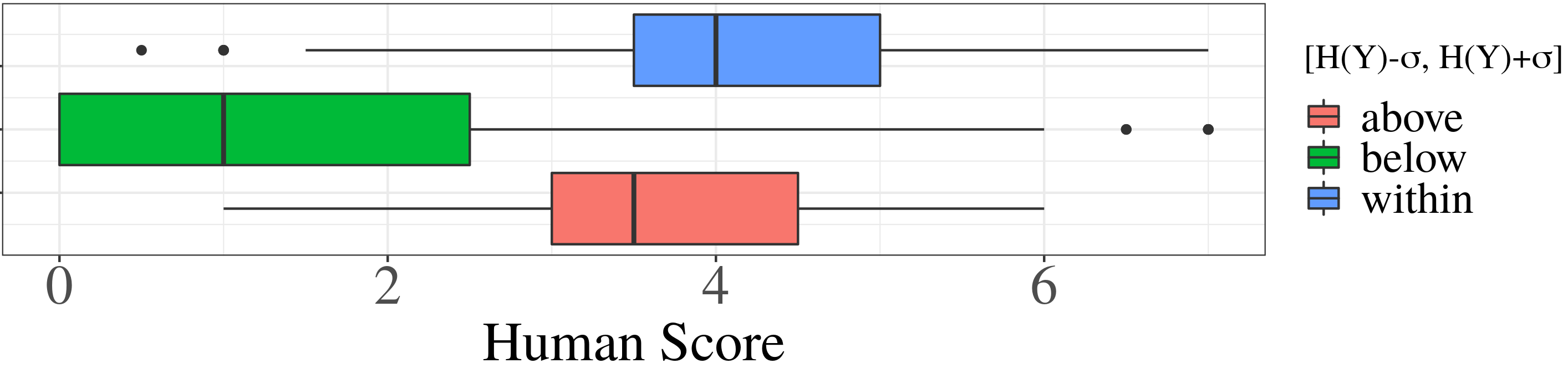}
    \caption{Human scores for strings (including both reference text and model-generated text) within 1 std of model entropy and above/below this interval. Note that ``above'' corresponds to text that has \emph{lower} probability than the specified interval; due to the nature of the decoding strategies explored in this work, which all  to some extent  (except for ancestral sampling) disproportionately favor higher probability strings, only $< 5\%$ of all strings evaluated fall into the ``above'' category. Thus, we do not have a representative evaluation of this region of the probability space. However, it is often observed that extremely low-probability strings are usually incoherent or nonsensical.}
    \label{fig:box_breakdown}
\end{figure}
\begin{figure}
    \centering
    \includegraphics[width=0.7\linewidth]{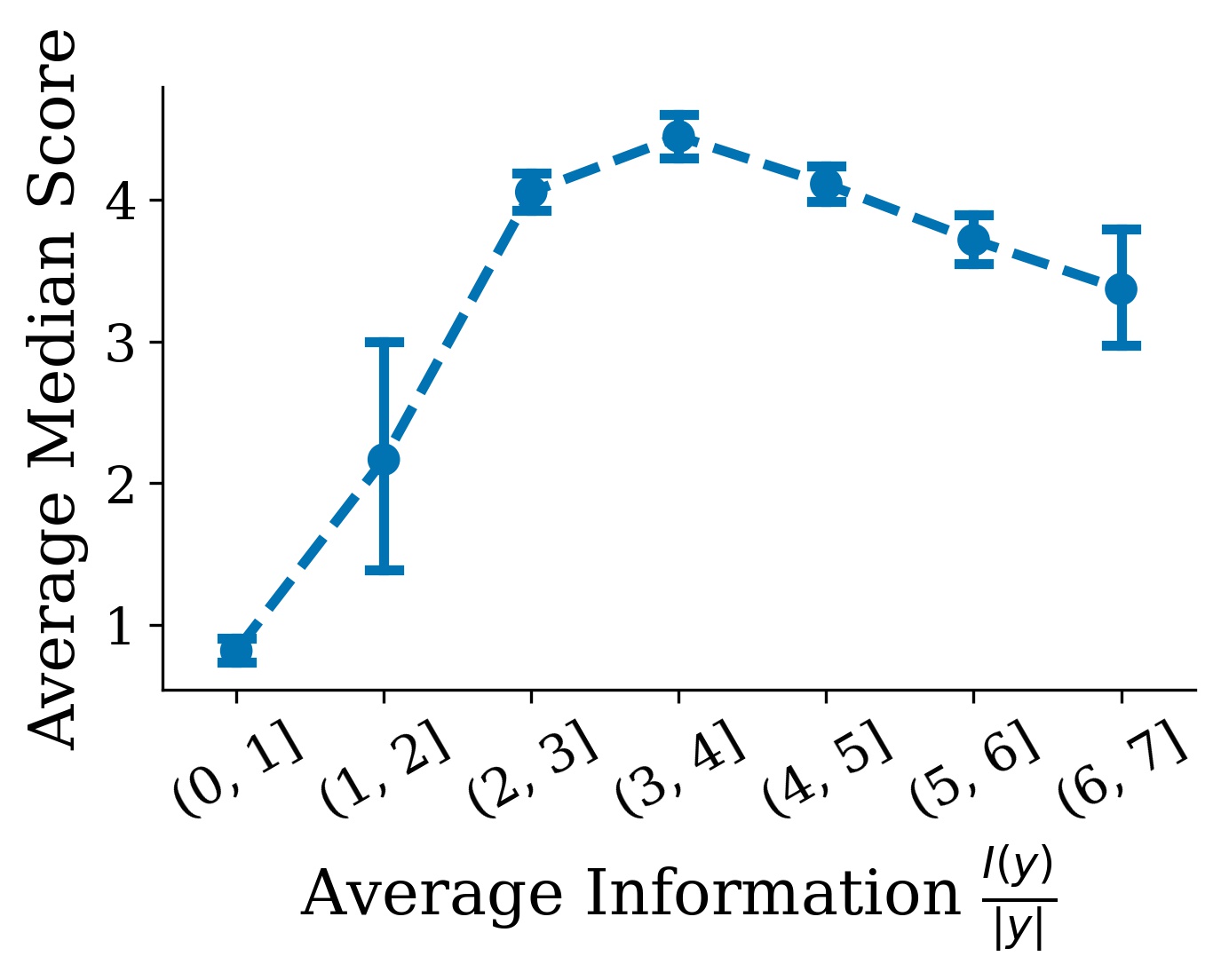}
    \caption{For story generation, median human scores (averaged across the two criterion) versus information, grouped by intervals; bars represent std. We normalize $\information(\yy)$ by length to mimic setup of  \citet{zhang_trading_2020}, which controls for length during generation. As with \citet{zhang_trading_2020}, we see an inflection point in the relationship along the information (equivalently, negative log-probability) axis.\looseness=-1}
    \label{fig:prob-v-qual}
\end{figure}

\begin{figure}
    \centering
    \includegraphics[width=\linewidth]{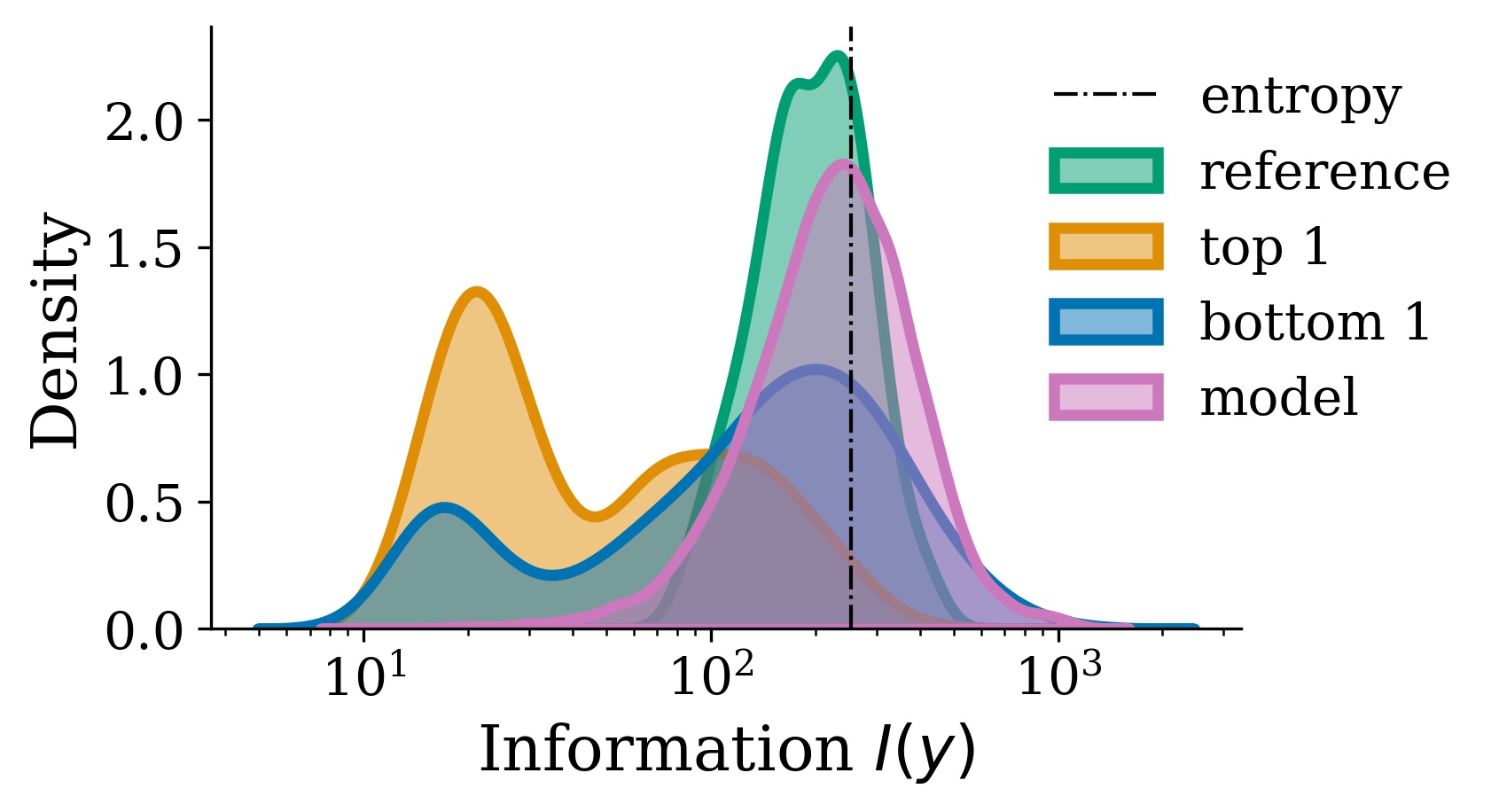}
    \caption{For abstractive summarization, the distribution over information $\information(\yy)$ values of: (model) the model, as estimated using samples from $\model$; (reference) the reference strings; model-generated strings ranked (top 1) first and (bottom 1) last among all decoding strategies by human annotators. The latter 3 are all w.r.t. a held-out test set. }
\end{figure}

\vspace{200pt}

\begin{figure}
    \centering
    \includegraphics[width=\linewidth]{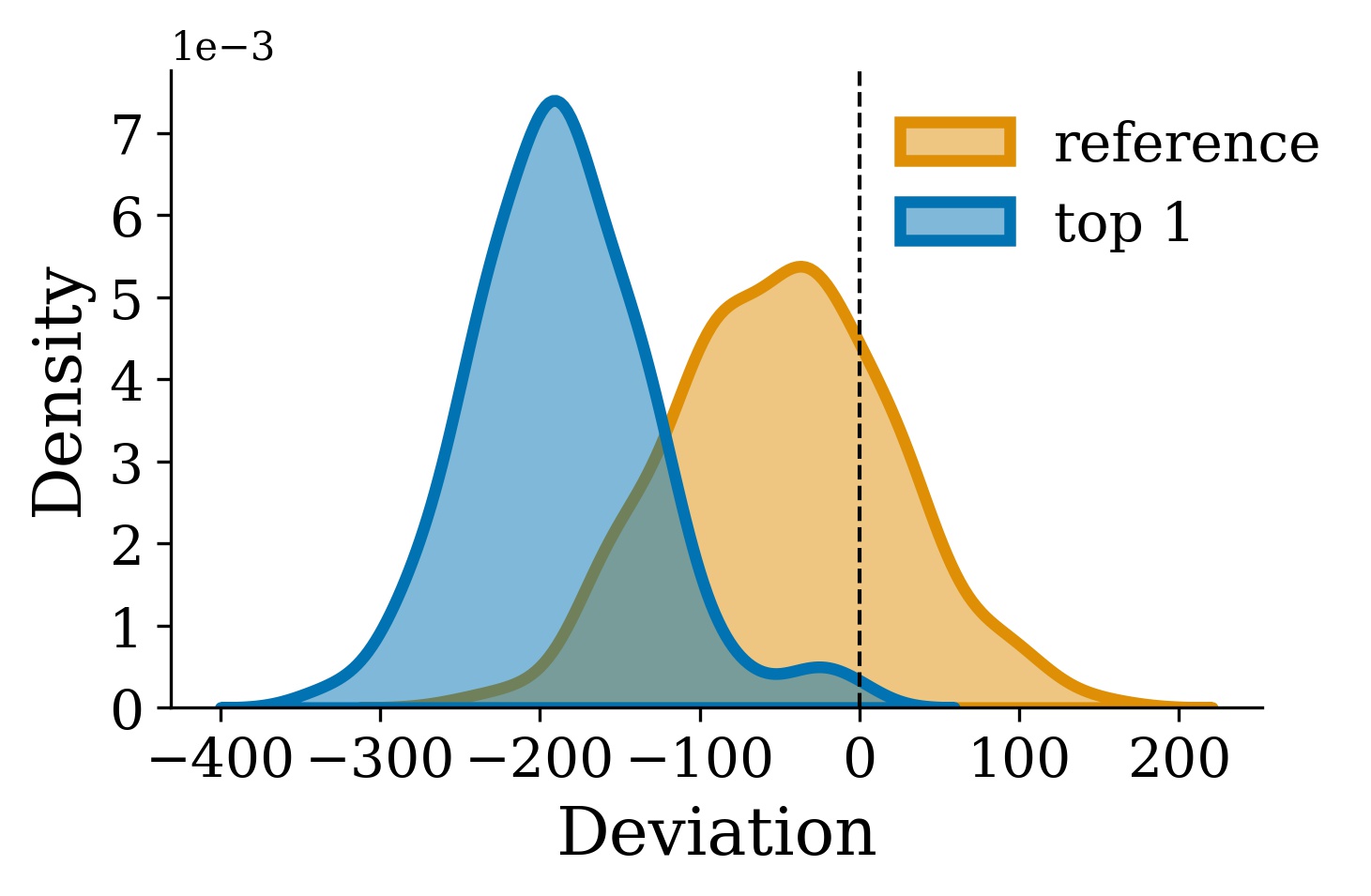}
    \caption{For abstractive summarization, the distribution of the difference in total information content for (1) test-set references and (2) top-ranked model-generated strings from the entropy of the model from which they were generated.}
\end{figure}

\begin{figure}
    \centering
    \includegraphics[width=\linewidth]{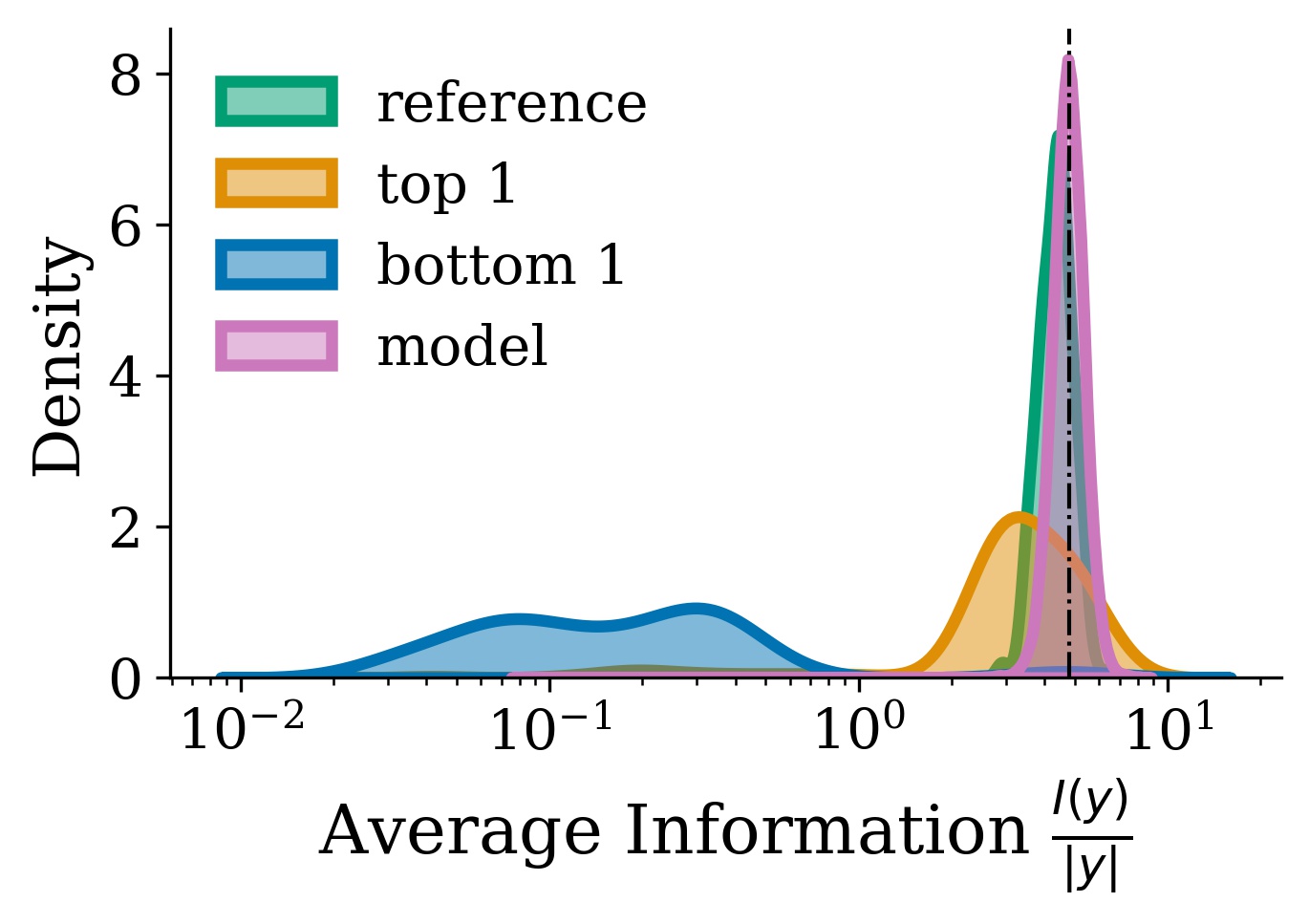}
    \caption{For story generation, the distribution over information ($
    \information(\yy)$) values normalized by length of: (model) the model, as estimated using samples from $\model$; (reference) the reference strings; model-generated strings ranked (top 1) first and (bottom 1) last among all decoding strategies by human annotators. The latter 3 are all w.r.t. a held-out test set. }
\end{figure}


\begin{figure*}
\centering
    \begin{subfigure}{\textwidth}
    \centering
    \includegraphics[height=0.35\linewidth]{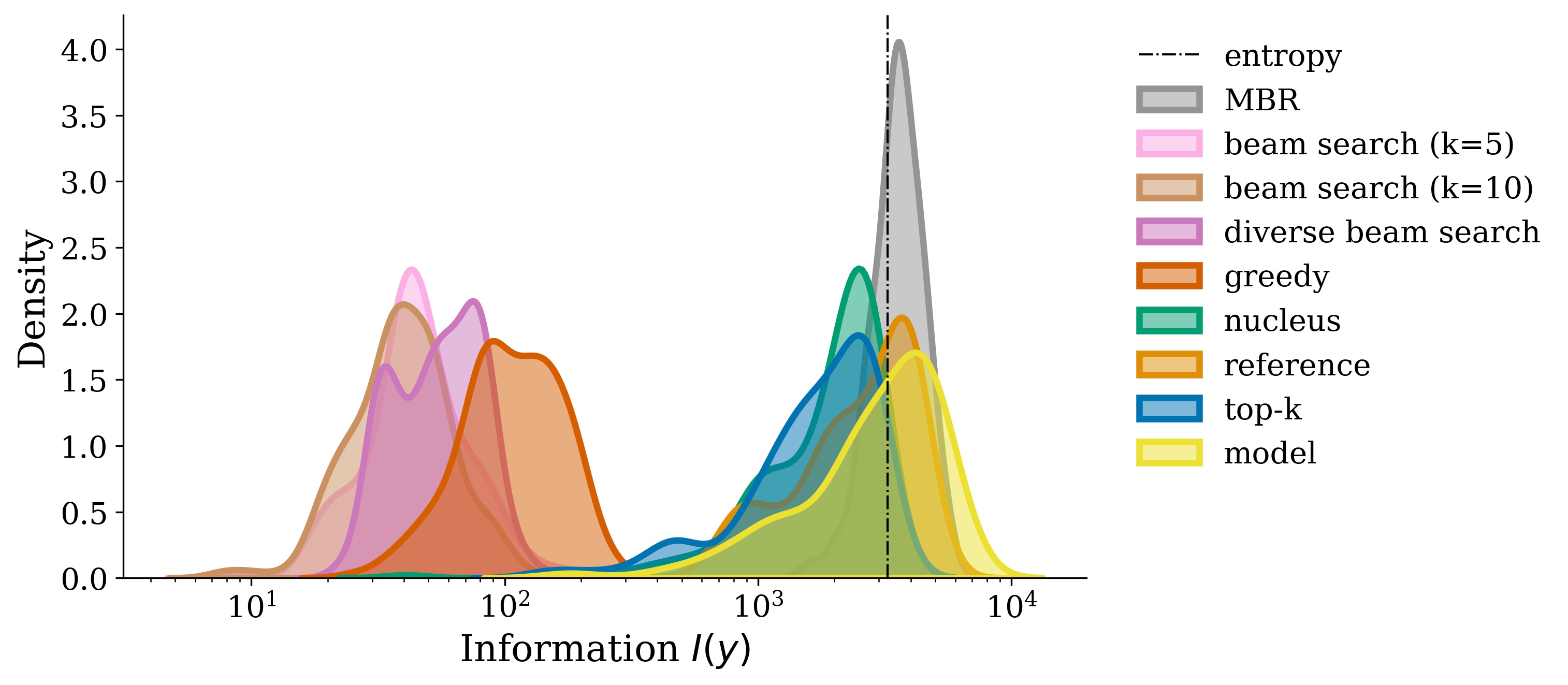}
    \caption{a}
    \label{test3}
\end{subfigure}
\hfill
\begin{subfigure}{\textwidth}
    \centering
    \includegraphics[height=0.38\linewidth]{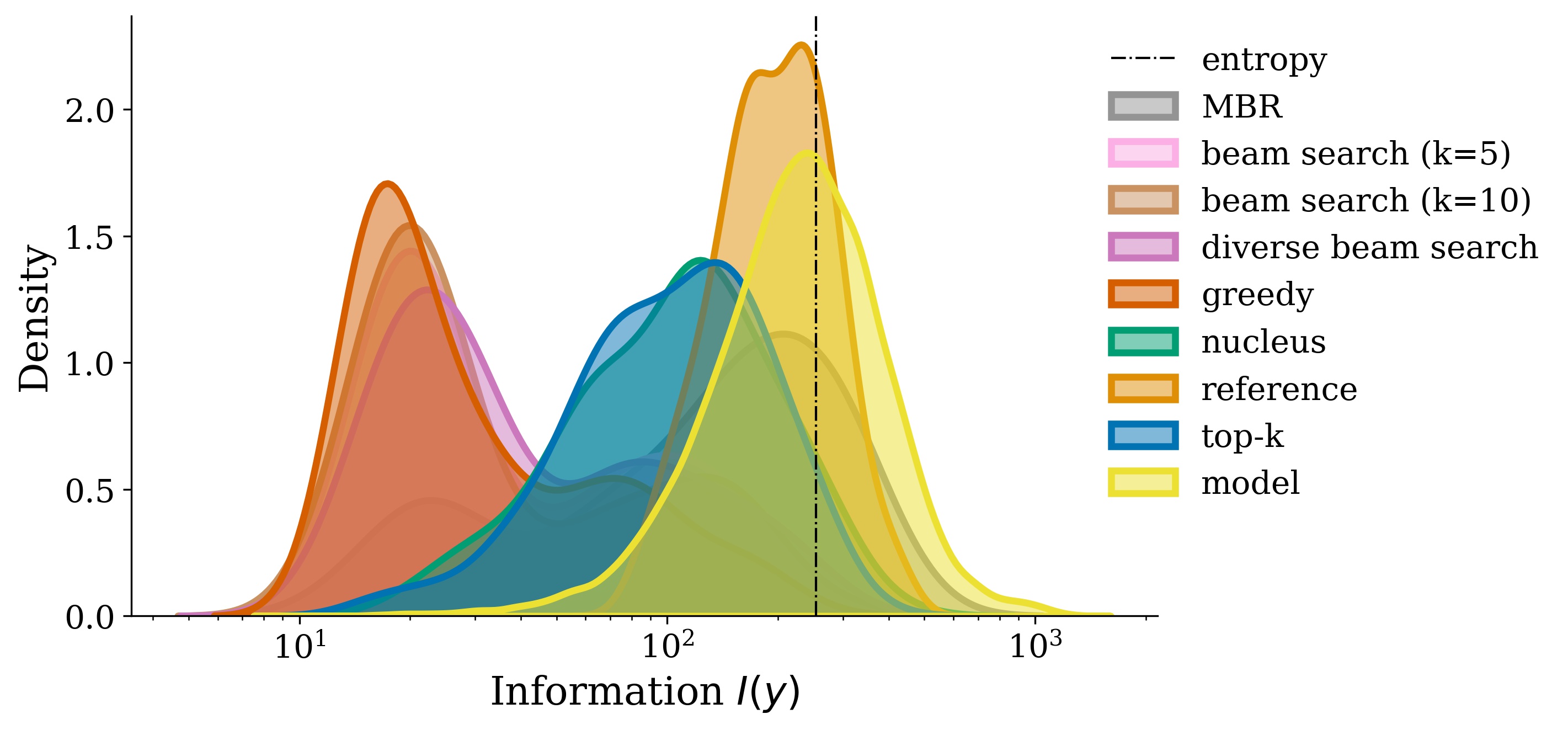}
    \caption{a}
    \label{test2}
\end{subfigure}
\caption{The distribution over information ($\information(\yy)$) values for strings generated under different decoding strategies for story generation (top) and abstractive summarization (bottom). Inputs are taken from a held-out test set. }
\end{figure*}

\end{document}